\documentclass[fleqn,10pt]{wlscirep}
\title{Adaptive compressed 3D imaging based on wavelet trees and Hadamard multiplexing with a single photon counting detector}

\author[1,3]{Huidong Dai}
\author[1,2,3]{Weiji He}
\author[1,*]{Guohua Gu}
\author[1]{Ling Ye}
\author[1]{Tianyi Mao}
\author[1]{Qian Chen}

\affil[1]{Jiangsu Key Lab of Spectral Imaging \& Intelligence Sense, Nanjing University of Science and Technology, Nanjing, 210094, China}
\affil[2]{Key Laboratory of Intelligent Perception and Systems for High-Dimensional Information of Ministry of Education, Nanjing University of Science and Technology, Nanjing, 210094, China}
\affil[3]{These authors contributed equally to this work.}
\affil[*]{gghnjust@mail.njust.edu.cn}

\begin{abstract}
Photon counting 3D imaging allows to obtain 3D images with single-photon sensitivity and sub-ns temporal resolution.
However, it is challenging to scale to high spatial resolution.
In this work, we demonstrate a photon counting 3D imaging technique with short-pulsed structured illumination and a single-pixel photon counting detector.
The proposed multi-resolution photon counting 3D imaging technique acquires a high-resolution 3D image from a coarse image and edges at successfully finer resolution sampled by Hadamard multiplexing along the wavelet trees.
The detected power is significantly increased thanks to the Hadamard multiplexing.
Both the required measurements and the reconstruction time can be significantly reduced by performing wavelet-tree-based regions of edges predication and Hadamard demultiplexing, which makes the proposed technique suitable for scenes with high spatial resolution.
The experimental results indicate that a 3D image at resolution up to $512\times512$  pixels can be acquired and retrieved with practical time as low as 17 seconds.  
\end{abstract}
\begin{document}

\flushbottom
\maketitle

\thispagestyle{empty}

\section*{Introduction}

Photon counting 3D imaging exploits pulsed illumination and photon counting detectors to acquire reflectivity and 3D structure of a scene at low light level.
Due to the single-photon sensitivity and sub-ns temporal resolution, it has drawn much attentions, ranging from biological imaging \cite{Becker2004Fluorescence,Kumar2007Multifocal} to long-range remote sensing \cite{Priedhorsky1996Laser,Herzfeld2014Algorithm}.

	The photon counting 3D imaging systems are categorized into two groups in terms of the number of detector elements, namely a single photon counting detector and photon counting detector arrays.
While they achieve high temporal resolution based on TOF principle \cite{Barbosa2010Remapping}, both of the two groups are challenging to scale to high spatial resolution.
Systems that scans through the scene pixel-by-pixel with a single photon counting detector \cite{Barbosa2010Remapping,Lamb2010Single,Mccarthy2009Long} is prevalent.
However, it suffers from system instability and long scanning time due to the mechanism of raster scanning.
The others instead uses photon counting detector arrays, such as the APD arrays used in the Jigsaw system created at MIT Lincoln Labs \cite{marino2005jigsaw} and the SPAD camera used for ultrafast imaging \cite{Gariepy2015Single,Gariepy2016Detection}.
While the single element photon counting detectors are well developed, the detector arrays are still in their infancy, with only resolution of $128\times32$ pixels available commercially, which suffers from high dark count rates and pixel cross-talk.
    
	To deal with applications with high spatial resolution, many recent efforts focus on optical multiplexing in a single-pixel camera configuration \cite{Duarte2008Single,Wakin2007An}.
These systems sense the scene through a series of coded projections using a spatial light modulator (e.g. a digital micro-mirror device) and a single photon counting detector.
Then the measurements and the foreknown patterns for projection are combined to retrieve the spatial resolution via a demultiplexing algorithm, which shifts the resolving power from the sensor to the spatial light modulator.
As a traditional method for multiplexing, Hadamard multiplexing is widely used in 2D imaging \cite{Pratt1969Hadamard,Streeter2009Optical,Mizuno2016Hadamard,Sun2017A} due to its high gather efficiency and low computational complexity demultiplexing.
Mingjie Sun et al. \cite{Sun2016Single} presented a single-pixel 3D imaging system which employs Hadamard multiplexing to sample the time-varying intensity.
Another photon counting 3D imaging systems \cite{Howland2011Photon,Howell2012Compressive,Howland2013Photon} employ optical multiplexing with a set of random matrices, and the spatial resolution can be retrieved based on compressed sensing (CS) principles \cite{Donoho2006Compressed,candes2006robust, Candes2006Near}.
These CS-based methods achieve non-scanning range acquisition using a single-pixel photon counting detector with much fewer measurements, but suffers from a computational overhead due to iterative optimization involved demultiplexing.
The consumed time increases exponentially with the increasing of spatial resolution, which limits the application.
    
	Recently, a technique named adaptive compressed sampling (ACS) \cite{Averbuch2012Adaptive} has been reported to achieve compressed sensing with low computational complexity.
Different from CS, ACS directly samples the significant wavelet coefficients with the patterns that form the wavelet basis.
The image can be recovered by performing an inverse wavelet transform, avoiding the computational overhead of CS.
In our previous work, we proposed an improved version (adaptive compressed sampling based on extended wavelet trees, EWT-ACS) \cite{Dai2014Adaptive} and applied the idea of multi-resolution wavelet basis sampling to single-pixel video acquisition \cite{Dai2016Adaptive}, color imaging \cite{Yan2016Colored}, and photon counting 3D imaging \cite{Dai2016Adaptivecomressed}.
However, the multi-resolution wavelet basis sampling strategy suffers from low gather efficiency, limiting the performance in photon-limited imaging.
In addition, since the technique computes every image slice of the 3D cube to extract the depth map, the computational complexity has a dependency on the number of time bins.
    
	In this work, We have demonstrated a single-pixel photon counting 3D imaging technique to deal with applications with high resolution, called adaptive compressed 3D imaging (AC3DI), which combines the advantages of Hadamard multiplexing and ACS.
Hadamard multiplexing is exploited to increase the photon gather efficiency, improving the image quality in photon counting condition.
The number of Hadamard patterns can be significantly reduced by only sampling edges of regions based on wavelet trees.
Since both the intensity image and depth map can be retrieved by a linear inverse Hadamard transform instead of the computational intensive optimization problems performed in CS, the time consumed can also be reduced.
Moreover, following the idea of collapsing the 3D cube data to a modulated image \cite{Howland2013Photon}, the technique has no dependency on the number of time bins, accelerating the reconstruction for long range imaging scenarios.

\section*{Results}
\subsection*{Adaptive compressed 3D imaging (AC3DI)} 
\begin{figure}
\centering
\includegraphics[width=0.75\textwidth]{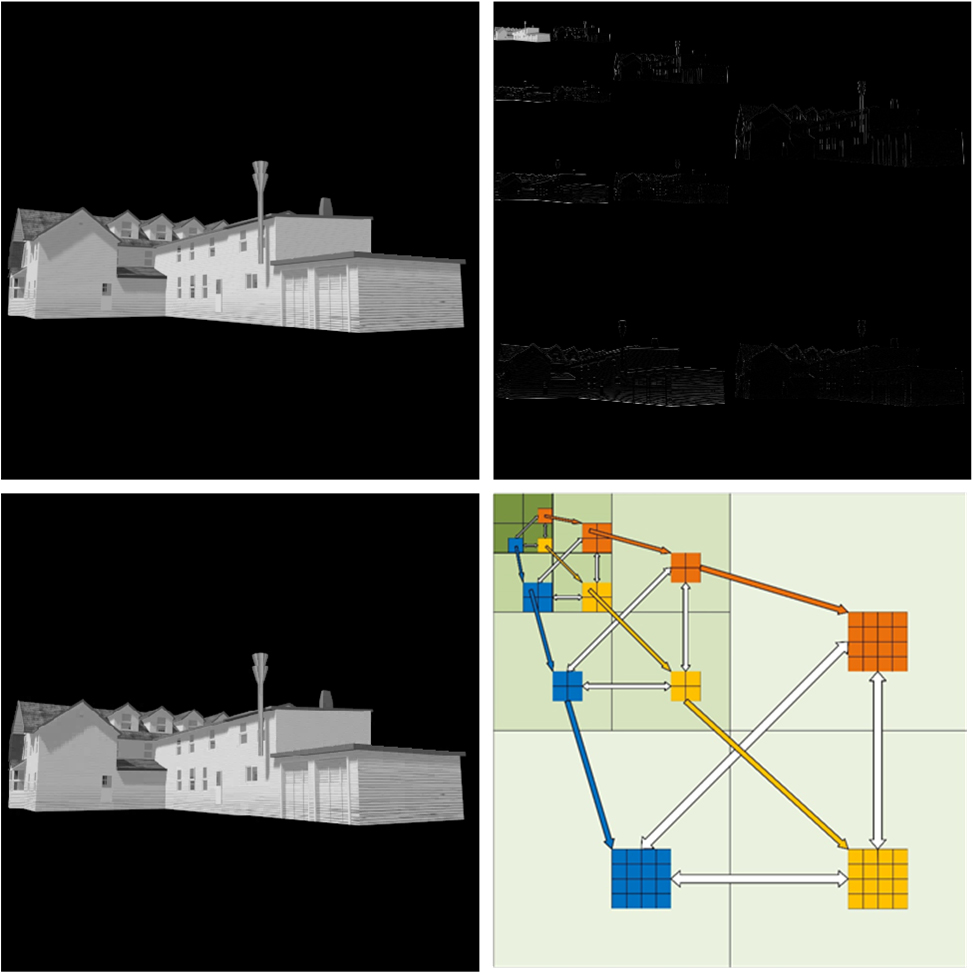}
\caption{
\textbf{Wavelet compression and the structure of extended wavelet trees.}
In the upper row, we show a $512\times512$ pixel intensity image (left) and its three-level wavelet representation in absolute values (right).
White regions represent significant coefficients corresponding to regions of sharp edges.
The image can be approximated without any regression (with a PSNR of 47.39 dB) using the largest 6.5 percent of the entire coefficients.
The wavelet transform provides a hierarchical modeling of wavelet tree structures over edges.
}
\label{fig:WCandWT}
\end{figure}

To better understand the AC3DI technique, we start with a brief introduction of the principle of wavelet compression, shown in Fig.\ref{fig:WCandWT}.
Being transformed to the wavelet domain, the wavelet representation is composed of scaling coefficients (the ones in upper left corner) and detail coefficients (the rest ones).
Wavelets can be regarded as multi-resolution edge detectors, where the scaling coefficients (the ones in upper left corner) represent a coarse version of the image and the absolution value of detail coefficients (the rest ones) represent the strength of the edges.
As shown in Fig.\ref{fig:WCandWT}, only small portion of detail coefficients corresponding to regions of edges, are large.
For the purpose of image compression, we can approximate an image without any regression by discarding most of the detail coefficients corresponding to regions of smoothness and only using the scaling coefficients and detail coefficients with absolute values larger than a predetermined threshold.

	However, the principle of wavelet compression can not be directly used for 3D image acquisition, because there are two issues remaining to be settled: the method of finding the regions of edges and the method of sampling 3D information.
On the first issue, the wavelet transform provides a hierarchical modeling of edges, i.e extended wavelet trees \cite{Dai2014Adaptive,Shapiro2002Embedded,Shapiro2002Embedded}, which is composed of the coefficients at different resolutions and different orientations but with the same spatial regions, shown in lower right of Fig.\ref{fig:WCandWT}.
It has been proven that if a detail coefficient corresponds to regions of edges, then with a very high probability its four children at the next higher resolution will also corresponds to regions of edges and needed to be sampled.
Therefore, the technique can predict the regions of edges to be sampled at the next resolution based on the wavelet representation  previously sampled.
On the second issue, we choose Hadamard multiplexing to sample the regions of edges.
Since a depth map cannot be straightforward recovered due to the nonlinearity between depth map and the acquired measurements, a modulated image is introduced to collapse the 3D cube data to 2D image. The depth map can be computed as the element-wise ratio of the modulated image and intensity image. 

    The idea behind AC3DI technique is to reduce the number of measurements needed to reconstruct the 3D image by only sampling a low resolution preview and regions of edges at various resolution with Hadamard multiplexing.
To that end, we first sample the scene with a low resolution set of Hadamard patterns, predict the regions of edges at the next higher resolution along wavelet trees, and only sample the union of regions of edges at higher resolution Hadamard multiplexing.
As for the remaining regions of smoothness, the resolution is enough to describe the scene, and no more patterns will be projected.
This process is repeated and the resolution improves by four times at each stage, until the final desired resolution is arrived. 

\begin{figure}
\centering
\includegraphics[width=0.75\textwidth]{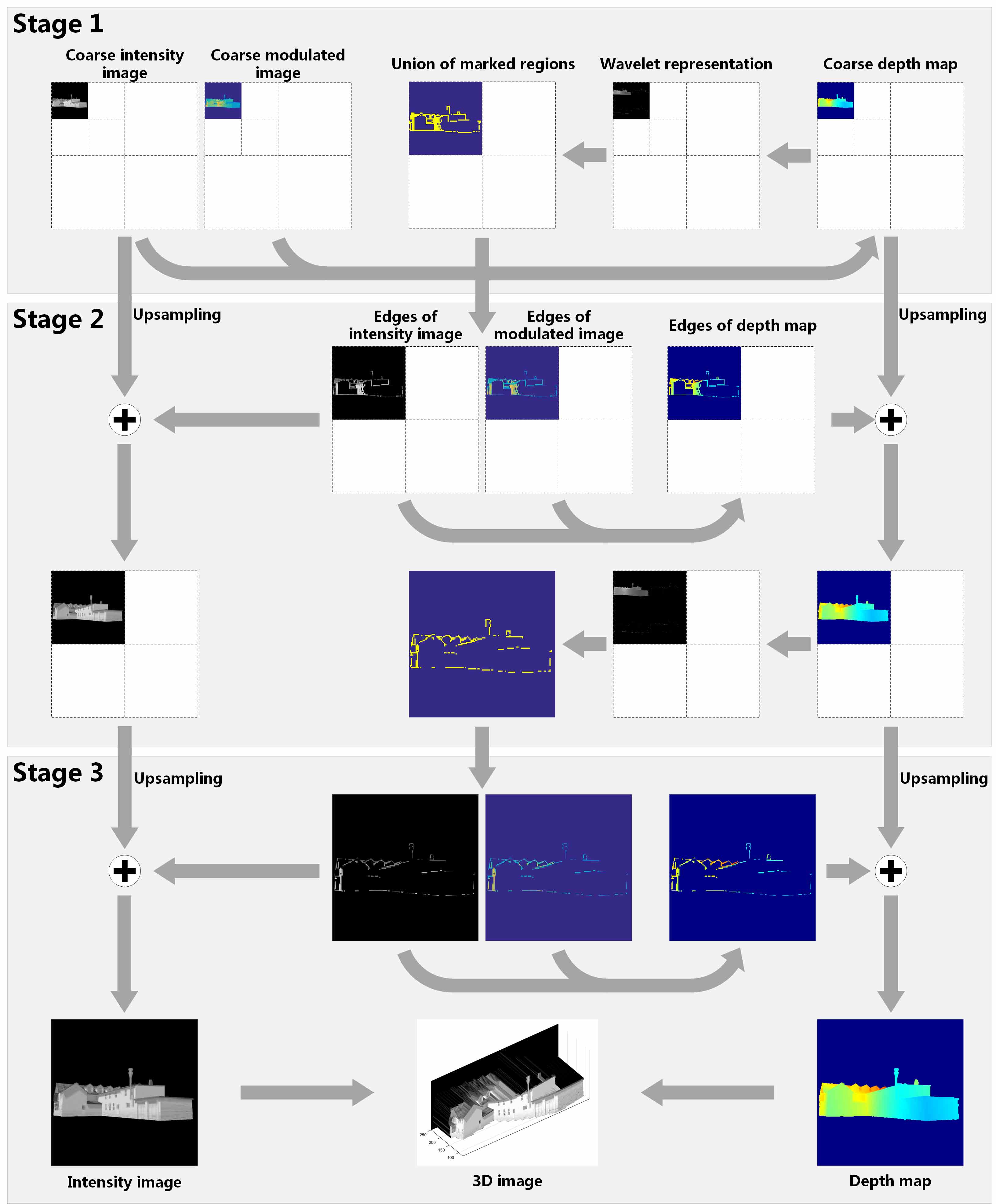}
\caption{
\textbf{The scheme of adaptive compressed 3D imaging.}
The object is a $512\times512$ pixel model of a house built by 3D Studio Max software.
In stage 1, we first acquire the $128\times128$ pixel intensity image and modulated image, and compute the corresponding coarse depth map. Regions of edges at the next higher resolution are predicted based on the wavelet representation of coarse depth map and marked as a union.
In stage 2, only the union of regions of edges is sampled by $256\times256$ pixel Hadamard multiplexing.
We upsample the intensity image and depth map at the last resolution, and combine them with the sampled edges to acquire $256\times256$ pixel intensity image and depth map.
In stage 3, the resolution reach $512\times512$ pixels.
The intensity image and depth map are perfectly retrieved, with PSNR of 31.63 dB and 34.28 dB, respectively.
Then, the 3D imaging can be acquired by combining the intensity image and depth map.
The total number of measurement to reconstruct the scene is only 10\% of the $512^2$ measurements in Nyquist sampling theorem.
}
\label{fig:AC3DI}
\end{figure}

    As an illumination of the AC3DI algorithm, we present a simulation on a $512\times512$ pixel model of a house built by 3D Studio Max software, shown in Fig. \ref{fig:AC3DI}.
In the first stage, the sampling process is initialized by projecting a set of $128\times128$ pixel Hadamard patterns onto the scene to acquire the coarse version of intensity image and modulated image.
The coarse version of depth map can be computed according to the modulated image and intensity image.
Then we perform a one-level wavelet transform on the coarse depth map, search the wavelet representation to find coefficients with absolute values larger than a predetermined threshold, and mark the corresponding regions.
The remaining regions corresponding coefficients with absolute values below the threshold will be discarded on the next stage, in order to reduce the number of patterns to be projected.
In the second stage, we first generate a set of $256\times256$ pixel patterns by sequential rearranging the Hadamard matrix into the union of the marked regions and remaining the other regions with zeros.
The generated patterns are loaded on the digital micro-mirror device (DMD) to sample the regions of edges at higher resolution.
The $256\times256$ pixel intensity image and depth map can be retried by combining the sampled edges and upsampling versions of the $128\times128$ pixel intensity image and modulated image, respectively.
Then technique computes the $256\times256$ pixel depth map, repeats the one-level wavelet transformation in order to search again for regions of edges at higher resolution along wavelet trees, and marks them.
As the algorithm goes on, the resolution of intensity image and depth map improves by four times for each stage. Finally, a $512\times512$ pixel 3D imaging can be acquired by combining the intensity image and depth map.

\subsection*{Experimental results}
\begin{figure}
\centering
\includegraphics[width=0.75\textwidth]{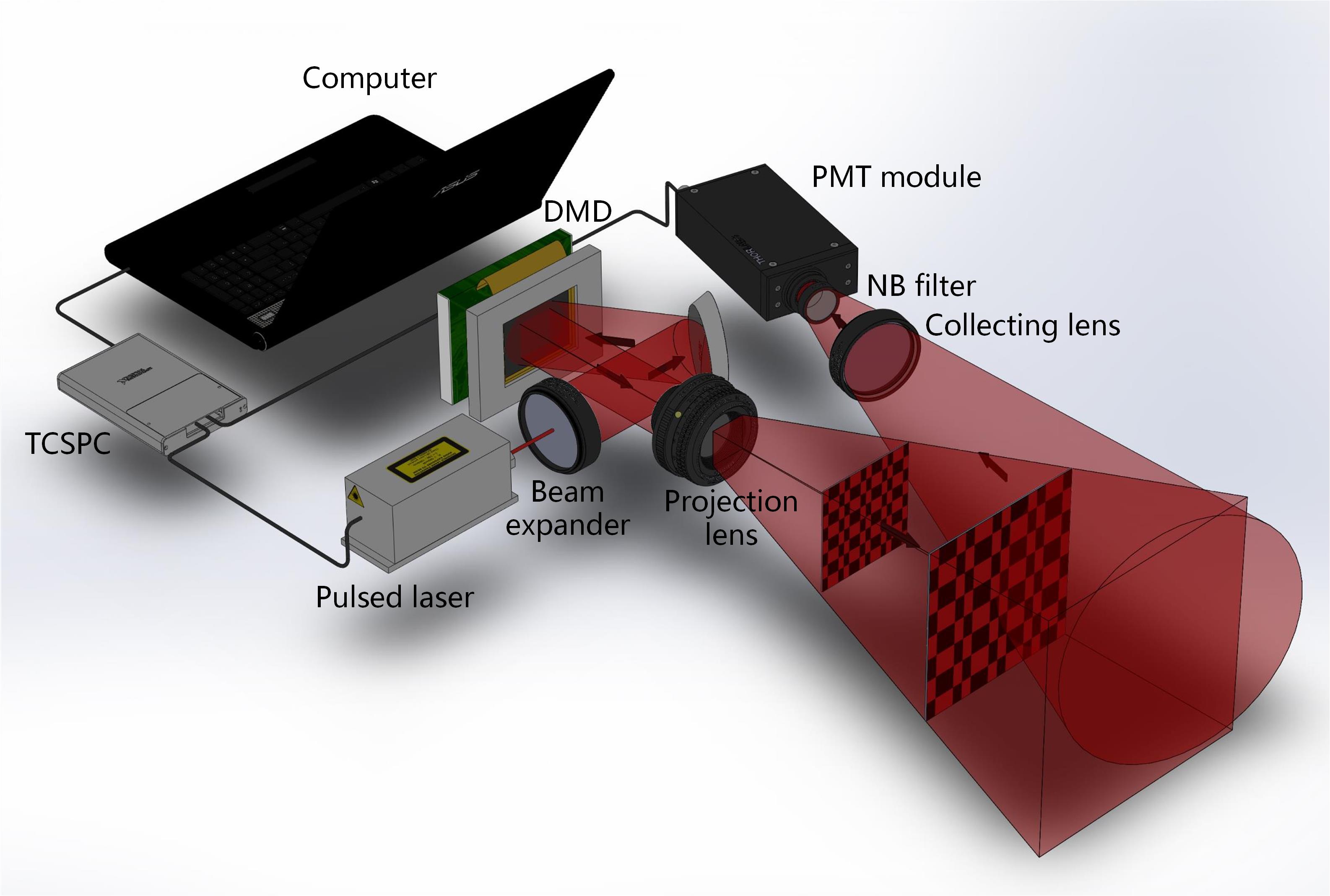}
\caption{
\textbf{The experimental setup used for AC3DI.}
A pulsed laser uniformly illuminates a DMD to project Hadamard patterns onto a scene, and the back-scattered photons are detected by a PMT module. The output signal is correlated with the transmitting pulses by a TCSPC module to find each photon's TOF, which can be used to reconstruct both intensity image and depth map of the scene by a demultiplexing algorithm.
}
\label{fig:Setup}
\end{figure}

In order to test the proposed technique, we demonstrate a 3D imaging system in the single-pixel camera configuration.  
The single-pixel camera configuration can be adapted for 3D imaging by using an active pulsed laser and a photon counting detection element together with a time-correlated single-photon counting (TCSPC) module, as shown in Fig.\ref{fig:Setup}.
The combination of a pulsed laser and a digital micro-mirror device (DMD) is implemented to provide structured illumination onto a scene.
The back-scattered photons are measured by a photomultiplier tube (PMT) module, and then be correlated with the transmitting pluses to find each photon’s TOF with the TCSPC module.
The optical multiplexing is performed by projecting the patterns displayed on the digital micro-mirror device (DMD), and therefore the intensity image and depth map can be retrieved by the following demultiplexing algorithm.  

\begin{figure}
\centering
\includegraphics[width=0.75\textwidth]{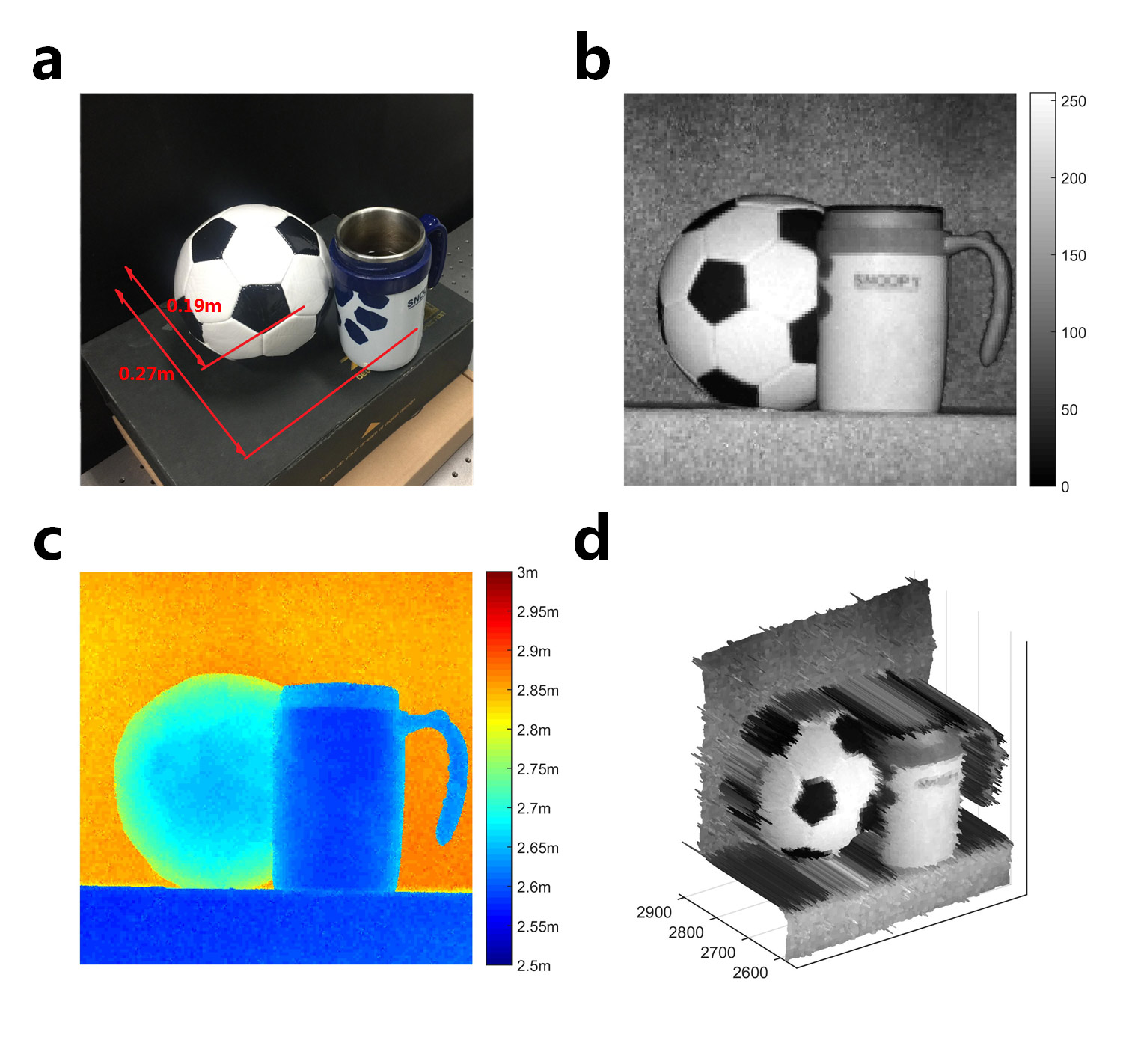}
\caption{
\textbf{3D imaging of a scene.}
(\textbf{a}) The photograph of the scene containing multiple objects.
The $512\times512$ pixel intensity image (\textbf{b}) and depth map (\textbf{c}) of the scene reconstructed with 15\% of measurements of the total number of pixels.
(\textbf{d}) The reconstructed 3D image of the scene.
}
\label{fig:Result1}
\end{figure}

In the experiment, a scene containing a 130 mm diameter football, a 140 mm tall mug, and a rubber cushion used as background was located within a distance of about 2.5-3 m, illuminated in Fig. \ref{fig:Result1}a.
We started the imaging with a $64\times64$ pixel coarse image and finally acquired the $512\times512$ pixel 3D image after four stages.
Both the intensity image and depth map were acquired with only 15\% of the total $512^2$ measurements in Nyquist sampling theorem, presented in Fig. \ref{fig:Result1}b,c.
The total time for acquisition, data transmission, and reconstruction were about 42 seconds, where the reconstruction time is less than 1 second.
A 3D image of the scene (shown in Fig. \ref{fig:Result1}d) was obtained by combining the intensity image and depth map, where the profile and some details of the objects could be distinguished.

\begin{figure}
\centering
\includegraphics[width=0.75\textwidth]{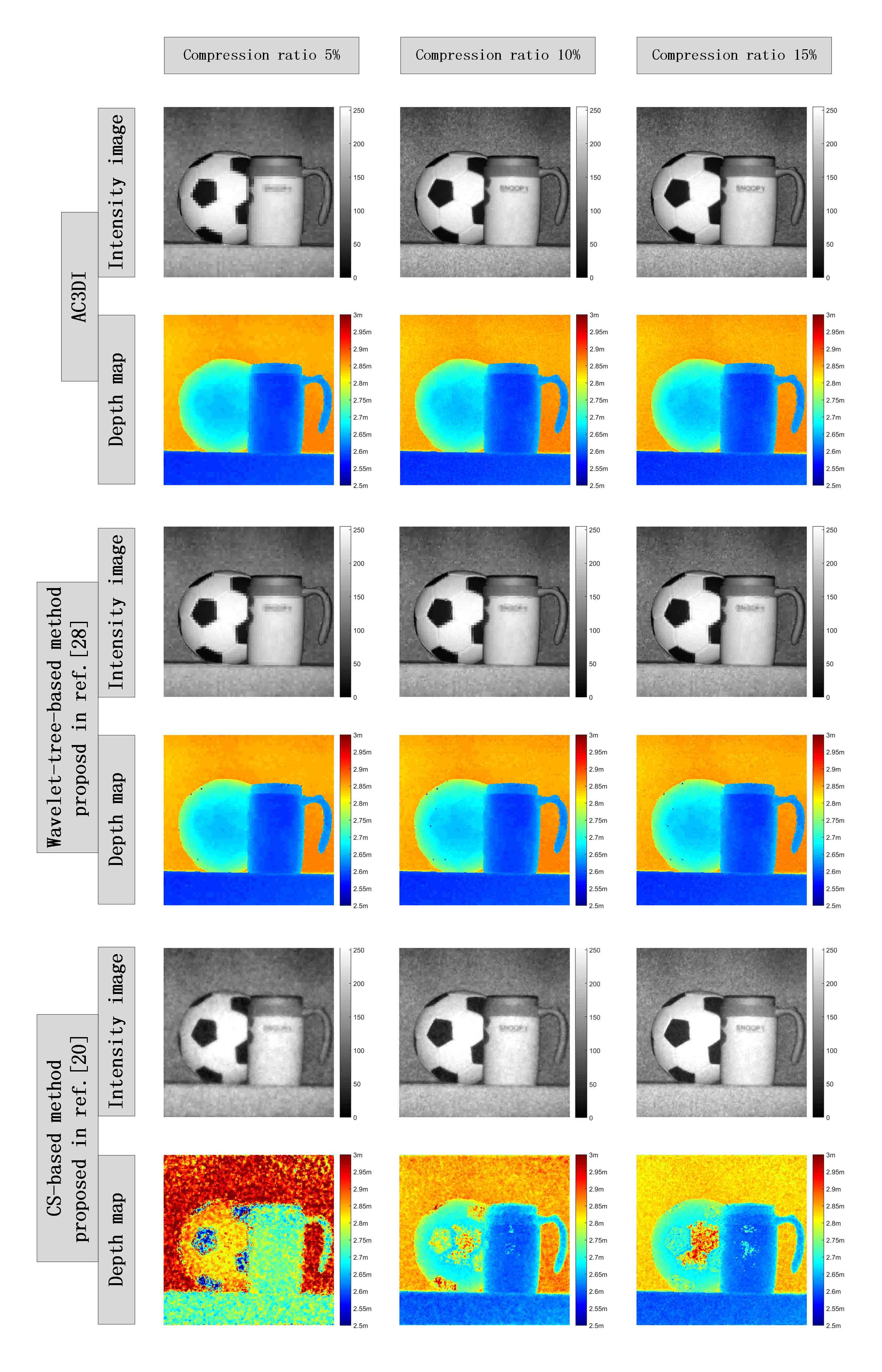}
\caption{
\textbf{Compression between our technique and the methods proposed in ref.[\cite{Dai2016Adaptivecomressed}] and ref.[\cite{Howland2013Photon}].}
The $512\times512$ pixel intensity images and depth maps retrieved by the AC3DI method, the wavelet-tree-based method proposed in ref.[\cite{Dai2016Adaptivecomressed}], and the CS-based method proposed in ref.[\cite{Howland2013Photon}] with compression ratios of 5\%, 10\%, and 15\% were presented. 
}
\label{fig:Result2}
\end{figure}

To better present the performance of the AC3DI technique, we compared our technique with two of the state-of-the-art compressed 3D imaging methods, i.e the wavelet-tree-based method proposed in ref.[\cite{Dai2016Adaptivecomressed}] and the CS-based method proposed in ref.[\cite{Howland2013Photon}].
As the AC3DI technique, the former method also searches the scene along the wavelet trees, but directly samples the significant wavelet coefficients with the patterns that form the wavelet basis, instead of the Hadamard multiplexing. The idea behind the method is to adaptively scanning significant information in the wavelet domain, leading to a lower power gather efficiency.
In addition, the time for reconstruction has a dependence of the number of time bins where there are objects at the corresponding depths. 
The latter method introduces a modulated image to convert the depth to intensity for direct measurement and the reconstruction time has no dependence of the number of time bins. However, it suffers from the computational overhead of convex optimization algorithms in CS.
Using the same scene, 3D images reconstructed by the three methods at compression ratios of 5\%, 10\%, and 15\% are given in Fig. \ref{fig:Result2}.
In the first two rows, both the intensity image and depth map can be acquired with high quality by the AC3DI method in 17seconds at a compression ratio of 5\%.
As the increasing of the number of measurements, the words on the mug can be seen clearly at a compression ratio of 15\%.
The intensity images retrieved by the wavelet-tree-based method proposed in ref.[\cite{Dai2016Adaptivecomressed}] has nearly the same quality as those retrieved by AC3DI.
However, there are several visible distortions in the depth maps, due to the omission of some significant coefficients in the sampling process.
In the last two rows, the intensity images reconstructed by the CS-based method proposed in ref.[\cite{Howland2013Photon}] have a fussy background, and the depth maps are not so good at such low compression ratios.
Since the the time for acquisition and data transmission of the three methods are almost the same, we only presented the reconstruction time in Table.\ref{tab:Comparison} for comparison.
As can be seen, the AC3DI performs slightly better than the wavelet-tree-based method proposed in ref.[\cite{Dai2016Adaptivecomressed}] in computational time, with both of them less than 1 second.
Time consumed by the CS-based method proposed in ref.[\cite{Howland2013Photon}] are around 30 seconds, due to the computationally intensive reconstruction process in CS.

\begin{table}
\centering
\begin{tabular}{|c|c|c|c|}
\hline
Compression ratio & AC3DI & Method in ref.[28]  & Method in ref.[20] \\
\hline
5\%               & 0.64 & 0.83 & 31.86 \\
\hline
10\%              & 0.58 & 0.85 & 30.84 \\
\hline
15\%              & 0.59 & 0.90 & 39.48 \\
\hline
\end{tabular}
\caption{\label{tab:Comparison}
\textbf{Reconstruction time comparison between the AC3DI method, the methods proposed in ref.[\cite{Howland2013Photon}] and ref.[\cite{Howland2013Photon}].}
}
\end{table}

\section*{Discussion}
We have demonstrated an adaptive compressed 3D imaging (AC3DI) technique with short-pulsed structured illumination and a single-pixel photon counting detector.
The AC3DI technique exploits the sparsity of nature scenes in the wavelet domain, and only samples a coarse image and edges of regions at successfully higher resolutions by Hadamard multiplexing along wavelet trees, which significantly reduces the number of measurements needed to reconstruct the 3D image.
Furthermore, the AC3DI technique retrieves the intensity images and depth maps by a linear inverse Hadamard transform, avoiding the computational overhead in traditional compressed sensing.
The experimental results show that a $512\times512$ pixel 3D image can be acquired in 17 seconds with a compression ratio as low as 5\%.
In the comparison experiments, the AC3DI technique outperforms the other methods with regards to image quality and computational time.
It is worth mentioning that the AC3DI can be extended to infrared wavebands by employing an infrared pulsed laser and an InGaAs detector, benefit from the flexibility of the single-pixel camera configuration.
The atmospheric scattering is much weaker in infrared wavelength, which can  improve the imaging distance.

\section*{Methods}

\subsection*{Hadamard multiplexing}
Hadamard multiplexing is the technique of measuring groups of samples according to patterns taken from the Hadamard matrix $H$, which takes value of $\{+1,-1\}$ and with rows and columns that are orthogonal. The lowest-order Hadamard matrix is of order two, which is
\begin{equation}
\hspace*{\fill}
H = \begin{bmatrix} 1&1\\1&-1 \end{bmatrix}.
\hspace*{\fill}
\end{equation}
Higher order Hadamard matrices can be iteratively generated by
\begin{equation}
\hspace*{\fill}
H_{2^n} = H_{2^{n-1}} \otimes H_2,
\hspace*{\fill}
\label{eq:HGen}
\end{equation}
where $\otimes$ is the Kronecker product operator. 

	Note that the entries of a Hadamard matrix are ‘+1’ or ‘-1’, which are not directly applicable to DMD.
In this work, we circumvent this by shifting the ‘-1’ entries to ‘0’.
Each row from the zero-shifted Hadamard matrix is reshaped into a 2D pattern to provide structure illumination.
Then, the measurements are subtracted by an appropriate mean in post processing to boost the demultiplexing. Hadamard multiplexing has the following advantages:
(a) All rows of the Hadamard matrix are orthogonal to each other, which allows us to obtain the image with less patterns.
(b) The number of entries ‘1’ and ‘0’ are equal in zero-shifted Hadamard matrix, which provides structured illumination with a constant light-level, improving both the signal-to-noise ratio (SNR) and signal gathering efficiency.
(c) The generation and inversion of Hadamard matrices is linear, requiring low computational complexity. 

\subsection*{Data acquisition and image cube reconstruction}
	To obtain a high resolution 3D image, the acquisition process is divided in to $J$ stage, and the resolution improves by four times for each stage.
We illustrate the acquisition process at the $j$-th stage for example ($j \neq J$ for universality).

	Let us assume that the intensity image and depth map needed to be acquired is at resolution of $\sqrt{N} \times \sqrt{N}$ pixels in the $j$-th stage.
According to the $\sqrt{N}/2 \times \sqrt{N}/2$ pixel wavelet representation of depth map in the last stage ${W_D}'$, the technique searches for regions of edges at a higher resolution along wavelet trees in the $j$-th stage.
The results are shown in an $N \times 1$ pixel binary matrix $Mark$, where $Mark(n) = 1$ indicates the pixel $(p,q)$, where $ p = \lfloor n/\sqrt{N} \rfloor + 1 $ and $ q = n\bmod \sqrt{N} $, corresponds to edges and will be sampled then, while  $Mark(n) = 0$ indicates $(p,q)$ contains no edges at the higher resolution and no more patterns are needed. Then the predication process performs as
\begin{equation}
\hspace*{\fill}
Mark(n) = 
\begin{cases}
1 & \text{if} \left|{\bar{W}_D(p,q)} \right| >= Threshold,\\
0 & \text{if} \left|{\bar{W}_D(p,q)} \right| < Threshold,
\end{cases}
\hspace*{\fill}
\label{eq:Mark}
\end{equation}
where $Threshold$ is a predetermined threshold to judge the judge the sharpness of edges in the corresponding regions, $\bar{W}_D = U({W_D}')$, and $U(\cdot)$ is the upsampling operator improving the resolution four times, such that
\begin{equation}
\hspace*{\fill}
U(X) = X \otimes \begin{bmatrix} 1&1\\1&1 \end{bmatrix}
\hspace*{\fill}
\label{eq:Upsampling}
\end{equation}

	According to the edge predication in the $(j-1)$-th stage, there are only $M = \sum_{n=1}^N Mark(n)$ pixels corresponding to the regions of edges that require to be sampled.
We generate a $L\times L$ pixel Hadamard matrix $H_L$ as Eq.\ref{eq:HGen} where $L = 2^{\lceil \log_2 M \rceil}$, and shift the entries '-1' to '0'.
Then the $M\times N$ sensing matrix $H$ in the $j$-th stage is generated by sequential rearranging the rows of zero-shifted Hadamard matrix $H_L$ into the regions of edges and remaining other regions with entries '0', such that
\begin{equation}
\hspace*{\fill}
H(m,n) = 
\begin{cases}
H_L(m,n_1) & \text{if} Mark(n) = 1,\\
0 & \text{if} Mark(n) = 0,
\end{cases}
\hspace*{\fill}
\label{eq:SMatGen}
\end{equation}
where $m$ is the row index of the sensing matrix indicating the index over patterns, $n$ is the column index of the sensing matrix indicating the index over DMD pixels, and $n_1=\sum_{nn=1}^n Mark(nn)$ denoting the index over pixels in regions of edges in the $m$-th pattern in the $j$-th stage.
Then the rows of the sensing matrix are reshaped to a set of patterns which are then placed sequentially on the DMD.

	An $M\times1$ measurement vector of the intensity edge image $Y_I$ is obtained by recording the number of detected photons for each pattern, such that
\begin{equation}
\hspace*{\fill}
Y_I = HX_I = \sum_{m=1}^{M} \sum_{n=1}^{N} h_{mn} \eta_{mn}
\hspace*{\fill}
\label{eq:Yi}
\end{equation}
where $X_I$ is the $N\times1$ vector of the intensity image, $h_{mn}$ is the element at $m$-th row and $n$-th column of measurement matrix indicating the DMD status (‘1’ or ‘0’) of the corresponding pattern, and $\eta _{mn}$ is the recorded number of photons reaching the $n$-th pixel during the $m$-th pattern.

	Since a depth map cannot be straightforward recovered due to the nonlinearity between depth map and the acquired measurements, a modulated image $X_Q$ which is made up of the element-wise product of the intensity image $X_I$ and depth map $X_D$ is introduced to collapse the 3D cube data to 2D image.
Then the $M\times1$ measurement vector of the modulated edge image $Y_Q$ is linearly sampled by summing the TOF of each photon detected for each pattern, such that
\begin{equation}
\hspace*{\fill}
Y_Q = HX_Q = C \sum_{m=1}^{M} \sum_{n=1}^{N} h_{mn} \eta_{mn} T_{mn}
\hspace*{\fill}
\label{eq:Yq}
\end{equation}
where $X_Q$ is the $N\times1$ vector of the modulated image, $T_{mn}$ is the TOF of a photon arriving at the $n$-th pixel during the $m$-th pattern, and $C$ is a constant factor converting photon number to intensity and TOF to depth.

	The edge intensity image $X_I^e$ can be retrieved by performing a linear Hadamard transform on $Y_I$. 
Then the technique upsamples the sampled intensity image in the $(j-1)$-th stage, and combines it with the retrieved edge intensity image ${\hat{X}_I}'$ in the $j$-th stage to obtain the retrieved intensity image $\hat{X}_I$, such that
\begin{equation}
\hspace*{\fill}
\hat{X}_I = iHT(Y_I) + U({\hat{X}_I}') \cdot (1-Mark)
\hspace*{\fill}
\label{eq:XiRec}
\end{equation}
where $iHT(\cdot)$ is the inverse Hadamard transform.
As the same way, we retrieve the edge modulated image as $iHT(Y_Q)$.
The depth map $X_D$ in the $j$-th stage can be recovered by
\begin{equation}
\hspace*{\fill}
X_D = Nz(a)\cdot iHT(Y_Q)\cdot/iHT(Y_I) + U({\hat{X}_D}') \cdot (1-Mark)
\hspace*{\fill}
\label{eq:XdRec}
\end{equation}
where $Nz(a)=1$ for nonzero $a$ and $0$ otherwise preventing the divide-by-zero situation, and ${\hat{X}_D}'$ is the sampled depth map in the $(j-1)$-th stage. 

	The technique moves to the next stage and carries on until the desired resolution is arrived.

\subsection*{Hardware specifications and operating configurations}
The following components were used in the experimental setup (Fig.\ref{fig:AC3DI}): a pulsed laser (PicoQuant LDH series, 830 nm center wavelength, 5mW average illumination power, 10MHz repetition rate, 300 ps full wideth at half maximum); a DMD (Texas Instruments Discovery 4100 DMD, $1024 \times 768$ pixels); a narrow-band filter (Thorlabs FL830-10); a projection lens (Nikon, $f$ = 50 mm); a collection lens (Nikon, $f$ = 25 mm); a Si photomultiplier tube (Hamamatsu photosensor module H7422P-50); and a TCSPC module (PicoQuant PicHarp 300).

	There are several important points worth mentioning. (a) The modulation rate of the DMD can be reach up to 22.7 kHz, however, the illumination time of each pattern is set to 1 ms to ensure that enough photons are detected. (b) The active area of the PMT is of 5 mm in diameter, used in conjunction with a 25 mm collection lens, giving a $11.4^{\circ}$ field-of-view. (c) The average count rate is 0.5 megacounts per second, and the corresponding average detected power is about 0.125 pW.

\bibliography{sample}
\section*{Acknowledgements}
This work was supported by the National Nature Science Foundation of China (Grant No. 61271332), the Fundamental Research Funds for the Central Universities (Grant No. 30920140112012), the Seventh Six-talent Peak Project of Jiangsu Province (Grant No. 2014-DZXX-007), the Fundamental Project for Low-light-level Night Vision Laboratory (Grant No. J20130501), and the Innovation Fundamental Project for Key laboratory of Intelligent Perception and Systems for High-Dimensional Information of Ministry of Education (Grant No. JYB201509).

\section*{Author contributions statement}
H.D. and W.H. conceived the idea of AC3DI. H.D., T.M. and L.Y. conducted the simulations and experiments. H.D. analysed the results and wrote the manuscript. All authors discussed the result and reviewed the manuscript.

\section*{Additional information}
\textbf{Competing financial interests:} The authors declare that they have no competing interests.

\end{document}